\documentclass[conference]{IEEEtran}
\IEEEoverridecommandlockouts
\usepackage{cite}
\usepackage{amsmath,amssymb,amsfonts}
\usepackage{algorithmic}
\usepackage{graphicx}
\usepackage{float}
\usepackage{textcomp}
\usepackage{url}
\usepackage{siunitx}
\usepackage{xcolor}
\usepackage{booktabs}
\usepackage[flushleft]{threeparttable}
\usepackage[hidelinks]{hyperref} 
\usepackage{lettrine} 
\hypersetup{colorlinks=true, linkcolor=blue, citecolor=blue } 

\usepackage{lipsum}
\def\BibTeX{{\rm B\kern-.05em{\sc i\kern-.025em b}\kern-.08em
    T\kern-.1667em\lower.7ex\hbox{E}\kern-.125emX}}

\begin{document}

\title{Deep Learning for In-Orbit Cloud Segmentation and Classification in Hyperspectral Satellite Data}

\author{
\IEEEauthorblockN{Daniel~Kovac\IEEEauthorrefmark{1},
                Jan~Mucha\IEEEauthorrefmark{1},
                Jon~Alvarez~Justo\IEEEauthorrefmark{3}, 
                Jiri~Mekyska\IEEEauthorrefmark{1},
                Zoltan~Galaz\IEEEauthorrefmark{1},
                Krystof~Novotny\IEEEauthorrefmark{1}
                Radoslav~Pitonak\IEEEauthorrefmark{2},\\
                Jan~Knezik\IEEEauthorrefmark{2}, 
                Jonas Herec\IEEEauthorrefmark{2},
                Tor~Arne~Johansen\IEEEauthorrefmark{3}}
									
\IEEEauthorblockA{\IEEEauthorrefmark{1}Dept. of Telecommunications, Brno University of Technology, 61600, Brno, Czech Republic}
\IEEEauthorblockA{\IEEEauthorrefmark{2} Zaitra s.r.o., 60200, Brno, Czech Republic}
\IEEEauthorblockA{\IEEEauthorrefmark{3}Dept. of Engineering Cybernetics, Norwegian University of Science and Technology, Trondheim, Norway}

\thanks{This work was supported by the grant of the Technology Agency of the Czech Republic FW09020069 (On-board satellite AI-based system for effective hyperspectral data filtration and processing).}}

\maketitle


\begin{abstract}

This article explores the latest Convolutional Neural Networks (CNNs) for cloud detection aboard hyperspectral satellites. The performance of the latest 1D CNN (1D-Justo-LiuNet) and two recent 2D CNNs (nnU-net and 2D-Justo-UNet-Simple) for cloud segmentation and classification is assessed. Evaluation criteria include precision and computational efficiency for in-orbit deployment. Experiments utilize NASA's EO-1 Hyperion data, with varying spectral channel numbers after Principal Component Analysis. Results indicate that 1D-Justo-LiuNet achieves the highest accuracy, outperforming 2D CNNs, while maintaining compactness with larger spectral channel sets, albeit with increased inference times. However, the performance of 1D CNN degrades with significant channel reduction. In this context, the 2D-Justo-UNet-Simple offers the best balance for in-orbit deployment, considering precision, memory, and time costs. While nnU-net is suitable for on-ground processing, deployment of lightweight 1D-Justo-LiuNet is recommended for high-precision applications. Alternatively, lightweight 2D-Justo-UNet-Simple is recommended for balanced costs between timing and precision in orbit.


\end{abstract}

\begin{IEEEkeywords}
Hyperspectral Satellite Data, Cloud Segmentation, Classification, Convolutional Neural Networks, Principal Component Analysis
\end{IEEEkeywords}

\section{Introduction}

\lettrine[]{S}{pace missions} have historically incorporated RGB and multispectral imagers as part of their payloads for Earth observation. NASA/Landsat missions have employed traditional RGB imaging, while other missions such ESA/Sentinel-2B have utilized Multispectral Imaging (MSI). However, next-generation space missions are increasingly adopting Hyperspectral Imaging (HSI) over MSI due to its superior capability to detect features often inaccessible to traditional RGB and MSI acquisition methods. Missions like NTNU/HYPSO-1 and ASI/PRISMA exemplify the current utilization in orbit of hyperspectral (HS) instruments to measure and process light across a wide range of narrow spectral bands with high spectral resolution. The substantial amount of data generated by the existing HS missions, along with future HS missions like ESA/CHIME and NTNU/HYPSO-2, along with the increasing need for more sophisticated data analysis, can complicate the processing of HSI data when relying solely on conventional signal processing and machine learning techniques. Consequently, Deep Learning (DL) neural networks are rising as a promising solution for HSI data processing from simple to complex tasks. Some deep networks, such as Convolutional Neural Networks (CNNs), have demonstrated remarkable robustness in tasks like image feature extraction, due to their ability to learn and generalize effectively to non-linear data patterns. Nevertheless, the frequently high inherent complexity of these neural models for satellite missions has led to the utilization of commercial-off-the-shelf solutions including high-performance platforms like Field-Programmable Gate Arrays (FPGAs) to enable advanced on-board processing. These platforms, currently in use in HS space missions like HYPSO-1, enable the programmable deployment of neural models in orbit for various data processing tasks in flight such as semantic segmentation, while enabling power consumption to remain within the allocated power budget.




The literature within HSI segmentation has employed the use of deep neural networks, such as CNNs, for cloud detection in satellite HS imagery (see Table \ref{sota}). 
Recently in 2022, Jian et al. \cite{Jian2022} employed a transfer learning approach, incorporating the pre-trained networks VGG16 and ResNet50 into a U-Net architecture for cloud detection, which proved robust in scenarios with limited training HS acquisitions. Li et al. (2023) \cite{Li2023} introduced SAUNetCD, using spectral assimilation from different types of satellite images. This proved to be an effective solution for cloud detection across diverse imagery. Furthermore, Zhao et al. (2023) \cite{Zhao2023} introduced the Boundary-Aware Bilateral Fusion Network (BABFNet) to improve cloud detection, particularly focusing on challenging image regions like cloud boundaries and thin clouds. Moreover, Luo et al. (2022) \cite{Luo2022} introduced ECDNet, a lightweight CNN model optimizing computational efficiency while still focusing on achieving acceptable performance on datasets from Landsat 8 and MODIS for cloud detection. Grabowski et al. (2022) \cite{Grabowski2022b} presented the self-configurable nnU-Net for cloud detection with potential for on-board processing, demonstrating its superior performance without manual design. In a subsequent study in 2023 \cite{Grabowski2023}, the nnU-net underwent compression into a model of smaller size, which is convenient for in-orbit satellite deployment. The model's effectiveness was demonstrated in the competition On Cloud N: Cloud Cover Detection Challenge, achieving state-of-the-art performance. 


Given the critical need for computational efficiency in satellite operations highly constrained by limited power budgets, Pitonak et al. (2022) \cite{Pitonak2022} proposed the CloudSatNet-1. This hardware-accelerated quantized CNN, implemented on ground-based FPGAs, demonstrated high accuracy in cloud coverage segmentation. This enables, for instance, saving downlink resources through the reduction of cloudy data transmission. Moreover, recent works have also pursued further advancements in reducing the model size of CNNs for their deployment aboard satellites. For example, Justo et al. (2023) \cite{justo2023sea} introduced various lightweight 1D-CNN and 2D-CNN models trained on a dataset from the HYPSO-1 mission \cite{justo2023open} for the detection of clouds, oceans, and terrestrial features. The proposed lightweight networks not only surpassed the compression of state-of-the-art models but also outperformed them in segmentation performance. Justo et al. demonstrated that the proposed lightweight models, such as 1D-Justo-LiuNet and 2D-Justo-UNet-Simple, comparatively surpassed 14 additional models in performance and model size when utilizing all the HS bands. Remarkably, the authors demonstrated that their 1D CNNs outperformed 2D CNNs for HYPSO-1 HS imagery. 

The main objective of this study is to evaluate state-of-the-art CNN models and determine the most suitable one for cloud image classification with respect to accuracy, model size, inference time, and sensor independence, crucial factors for practical deployment on FPGAs.

\begin{table}[!tb]
\caption{Studies on Cloud Segmentation in Hyperspectral Imagery}\label{sota}
\centering
\resizebox{\columnwidth}{!}{%
\begin{tabular}{ccccc}
\hline
Study                & Year     & Method                    & Performance                \\ \hline
Zhao et al.  \cite{Zhao2023}           & 2023     & DFEM+BABFNet              & 97.7\% PA      \\
Li et al. \cite{Li2023}                & 2023     & SAUNetCD                  & 97.0\% PA       \\
Justo et al.  \cite{justo2023sea}      & 2023     & 2D-Justo-UNet-Simple           & 92.0\% PA  \\ 
Justo et al.  \cite{justo2023sea}      & 2023     & 1D-Justo-LiuNet           & 93.0\% PA  \\ 
Luo et al. \cite{Luo2022}              & 2022     & ECDNet                    & 0.93 CF1             \\ 
Grabowski et al.  \cite{Grabowski2022b}& 2022     & nnU-Net                   & 95.3\% PA      \\ 
Jian et al. \cite{Jian2022}            & 2022     & U-Net (VGG16+ResNet50)    & 91.0\% PA      \\
Pitonak et al. \cite{Pitonak2022}      & 2022     & CloudSatNet-1             & 88.2\% CA \\
\bottomrule
\multicolumn{4}{l}{\scriptsize{PA = Pixel Accuracy; CA = Classification Accuracy; CF1 = Classification F1 Score.}} \\
\end{tabular}%
}
\end{table}

\section{Dataset} \label{data}

\begin{figure}[!h]
\begin{center}

\includegraphics[width=0.85\columnwidth]{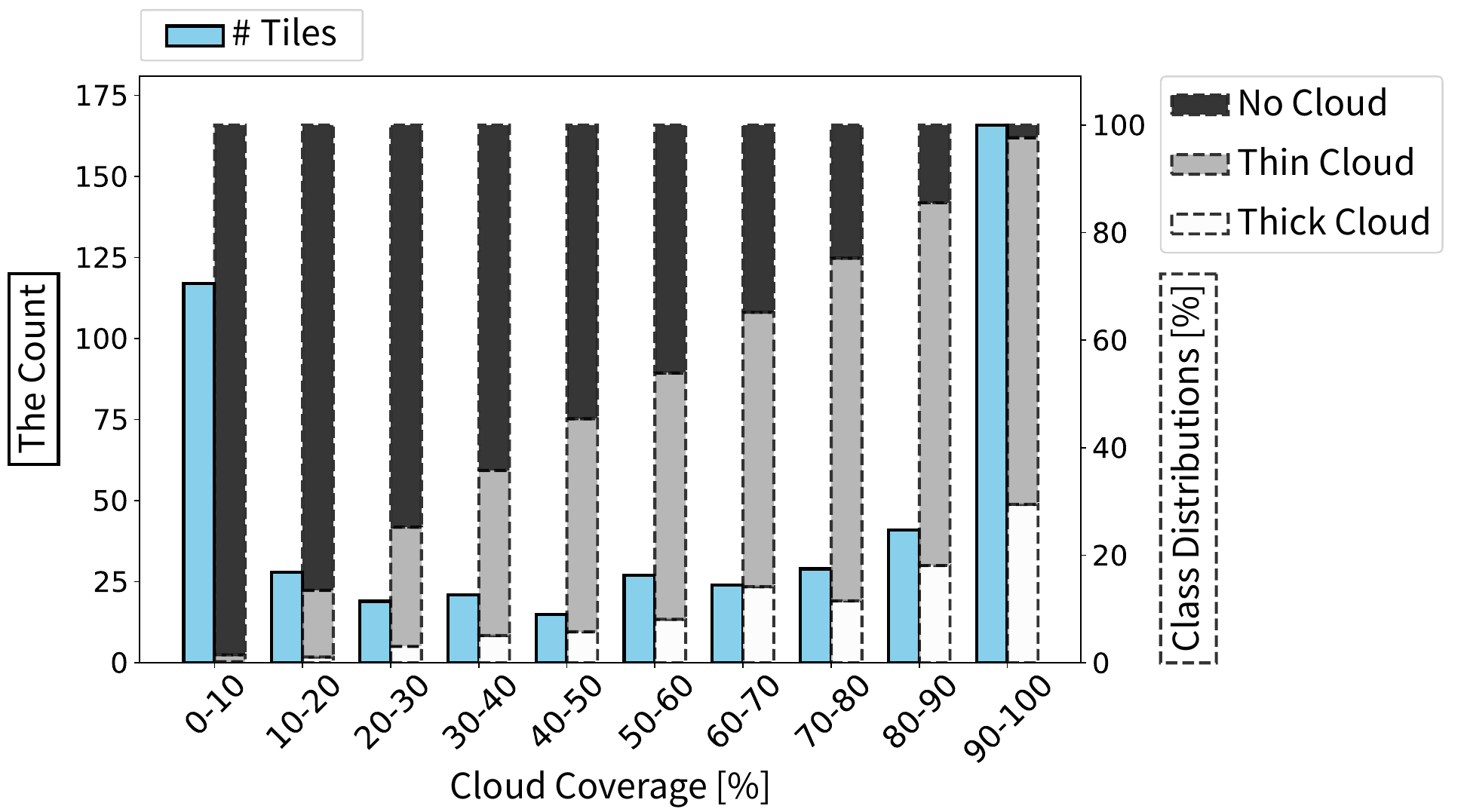}
\caption{Histogram Illustrating the Dataset's Tile Count (Left Y-axis) and Class Distribution (Right Y-axis) Relative to Cloud Coverage (X-axis).}\label{dataset}

\end{center}
\end{figure}

Our dataset consists of 35 HS images gathered from NASA's Earth Observing-1 (EO-1) mission. Designed as a technology demonstration mission with a nominal planned one-year duration, EO-1 defied expectations by remaining operational for nearly 17 years. Throughout its prolonged lifetime, EO-1 played a crucial role in numerous ecological and geophysical studies, mainly due to its powerful Hyperion sensor \cite{numata2011analyzing}. This sensor had an unprecedented combination of high spatial and spectral resolutions of 30 m and $\approx$ 10 \si{\nano\meter}, respectively, across 0.4 \si{\micro\meter} to 2.5 \si{\micro\meter} \cite{middleton2013earth}. Our dataset comprises the lowest data product level (L1R), with the primary emphasis on their relative proximity to unprocessed data directly accessible onboard, in accordance with the aim of future in-orbit deployment post-data acquisition. We choose calibrated VNIR and SWIR bands, comprising 198 out of 242 bands, which are then aligned and concatenated. Following this, we apply desmiling and destriping techniques, resulting in push-broom scenes with a width of 254 pixels and variable scanning lengths. These scenes are further divided into 254x254 px tiles for subsequent analysis. Given that the images lack labels for cloud detection needed for supervised learning, we adhere to the next annotation procedure. We first generate an RGB composite using bands 31 (661 \si{\nano\meter}), 21 (559 \si{\nano\meter}), and 13 (478 \si{\nano\meter}). To improve the composite's visual clarity, we add a fraction of the bands 123 (1380 \si{\nano\meter}) and 150 (1650 \si{\nano\meter}) to each RGB channel. These additional bands aid in distinguishing between low clouds, high clouds, and snow, as shown in Hyperion Cloud Cover \cite{griggin2003cloud}. Finally, we stretch linearly the RGB samples between the 1st and 97th percentiles to brighten the image. To achieve consensus among annotators, we begin by labeling a few images, then conduct a review and discussion to establish annotation guidelines for the rest of the dataset. Finally, we refine the expert-level annotations through a final review round. The resulting dataset consists of 487 tiles (254x254 px) with cloud masks including the following three categories: \textit{No Cloud}, \textit{Thin Cloud} and \textit{Thick Cloud}, with respective class distributions of 43.79 \%, 42.19 \%, and 14.02 \%. Fig. \ref{dataset} illustrates a histogram depicting the class distribution of the dataset. 


\section{Methodology}
\subsection{Principal Component Analysis} \label{pca}

Neural inference to predict the category for each pixel in an HS image can be computationally expensive due to the high spectral dimensionality of the images. Therefore, we employ Principal Component Analysis (PCA) to reduce the number of bands. By performing PCA exclusively on the ground segment, in-flight power can be conserved, saving some energy for subsequent cloud segmentation during satellite flight operations. Eliminating in-flight PCA processing can also aid in reducing latency. 
The primary objective of ground-based PCA is to identify channels holding essential information needed for satellite inference, as determined by their respective weights in the first principal component. Once the channels with the greatest variability are identified, direct channel selection can be used prior to in-flight inference. This method has been proposed and validated in prior studies \cite{justo2023sea}. 

\begin{figure*}[h!]
\begin{center}
\includegraphics[width=0.82\textwidth]{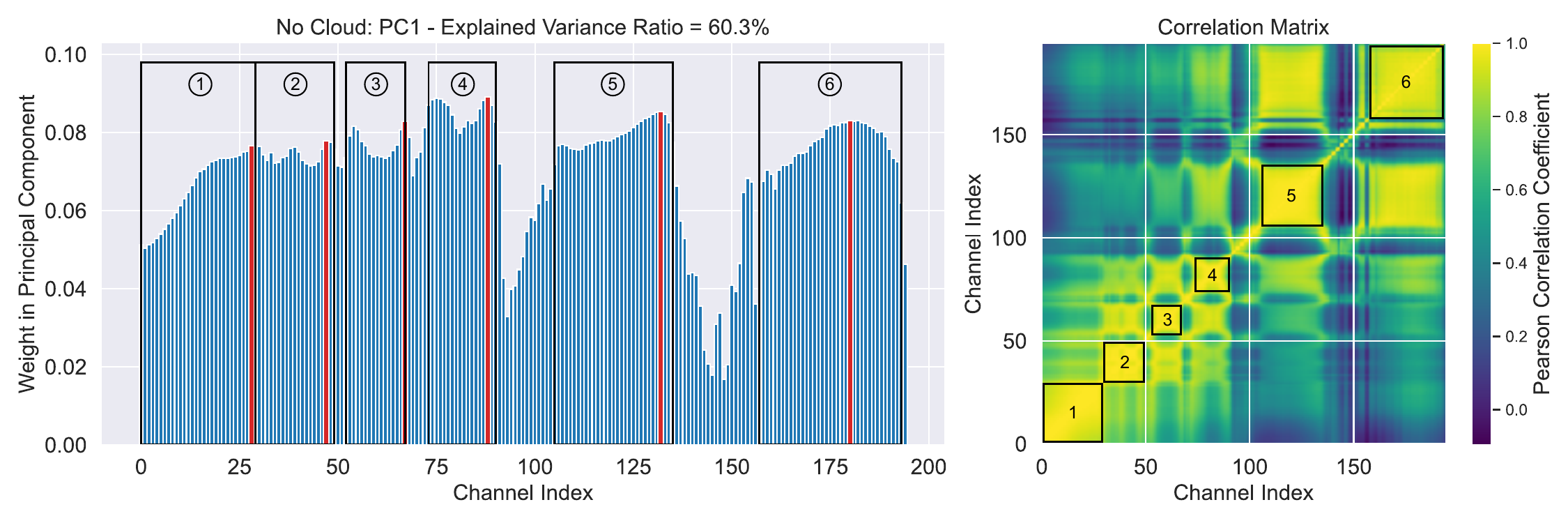}  
\caption{PCA and Correlation Matrix of the No Cloud Class: Identification of Correlation Clusters Highlighted in Black, with Emphasis on Peak Weights Highlighted in Red.}\label{pca_corr}
\end{center}
\end{figure*}

PCA is conducted on the EO-1 Hyperion dataset, after standardizing the data in each channel. We perform PCA in two scenarios. In the first one, we analyze all data and identify the channel with the highest weight on the principal component. This channel (index 77) corresponds to the 1205.07~\si{\nano\meter} wavelength, which falls in the SWIR band. In the second scenario, we perform PCA separately for the three classes \textit{No Cloud}, \textit{Thin Cloud}, and \textit{Thick Cloud}. Additionally, we employ correlation clustering, utilizing Pearson's correlation, to identify several clusters with the strongest correlation among channels. Next, we examine which channel exhibits the highest PCA weight associated with a particular cluster. Fig. \ref{pca_corr} demonstrates the identification of important channels for the \textit{No Cloud} class. In cases where significant channels vary within overlapping correlation clusters across classes, we prioritize selecting the one with the highest weight. This approach results in six important channels. The first (index 28) corresponds to a wavelength of 711.72 \si{\nano\meter} (visible red spectrum). Other channels (indexes 47, 67, and 88) at 905.05 \si{\nano\meter}, 1104.19 \si{\nano\meter}, and 1316.05 \si{\nano\meter}, respectively, capture NIR data. The last two channels (indexes 134 and 182) at 1759.89 \si{\nano\meter} and 2244.22 \si{\nano\meter} belong to the SWIR band.

\subsection{Deep Learning Modeling}

Based on the literature overview, we select three neural networks. First, nnU-Net \cite{isensee2021nnu}, a deep CNN for segmentation tasks featuring an automatic self-configuration mechanism, including pre-processing, network architecture, training, and post-processing. For the purpose of HS data processing, we use its 2D version. Next, we include the lightweight deep models 2D-Justo-UNet-Simple and 1D-Justo-LiuNet \cite{justo2023sea} recommended for in-orbit deployment due to their compact model sizes and high performance. The designations 2D and 1D signify their reliance on 2D CNN and 1D CNN architectures. 

For training the aforementioned models, we utilize the 487 image tiles from the EO-1 dataset, and partition them into training, validation, and test sets in a ratio of 70/20/10, respectively. Furthermore, for training the models we utilize a varying number of channels, falling under the following scenarios: one single channel (1205.07 \si{\nano\meter}) selected based on the highest weight in PCA across all classes combined; 6 channels determined by the PCA experiments described in the previous section; and 98 channels extracted from every second channel across the VNIR and SWIR spectra. In the latter scenario, we do not use all channels in the dataset due to the excessive RAM consumption this would pose for nnU-net given the self-configuring preprocessing before training. Inputting 98 channels into 1D-Justo-LiuNet enables the network to make predictions. However, since the network's four pooling layers gradually reduce spectral features by a subsampling factor of 2, when passing only 6 or 1 input channels, the features would rapidly diminish to 0, making predictions unfeasible. Thus, to test the configurations of 1 or 6 channels, we replicate channels within each tile: one channel is repeated 91x, while six channels are repeated 16x \cite{justo2023sea}. In the experiments involving the 2D-Justo-UNet-Simple, we segment each tile, initially sized at 254x254 px, into smaller overlapping tiles of dimensions 252x252 px. This adjustment is made to ensure that spatial dimensions remain integer after applying two pooling layers with a subsampling factor of 2 in the contracting path of the network. Subsequently, we train each model for 20 epochs, with 22 batch size. After the training phase, the inference employs a six-core (12 threads) Intel Core i5 12600 CPU with a 4.8 GHz maximum clock frequency.

During inference, we collect the predicted segmentation masks to evaluate the individual performance of the models using state-of-the-art metrics such as pixel-level accuracy and the Sørensen–Dice coefficient. Furthermore, for the classification of cloudy image tiles, we employ accuracy and F1 score. In our experiments, a tile is considered cloudy when the combined predicted coverage of the \textit{Thin Cloud} and \textit{Thick Cloud} classes exceeds 70 \%, following the recommendations from ESA's Phisat-1 mission \cite{cloudscout-vpu}. Finally, we further verify the models' generalization capability and its sensor independence using images from the commercially available Ziyuan-1 02 dataset \cite{wu2013data}, which has the exact spatial resolution of 30 \si{m} as EO-1 Hyperion and contains 166 sensor channels, also divided into VNIR and SWIR. Six channels matching wavelengths identified by PCA are chosen for model analysis.

\section{Results}

\begin{figure*}[!h]
\begin{center}
\includegraphics[width=0.91\textwidth]{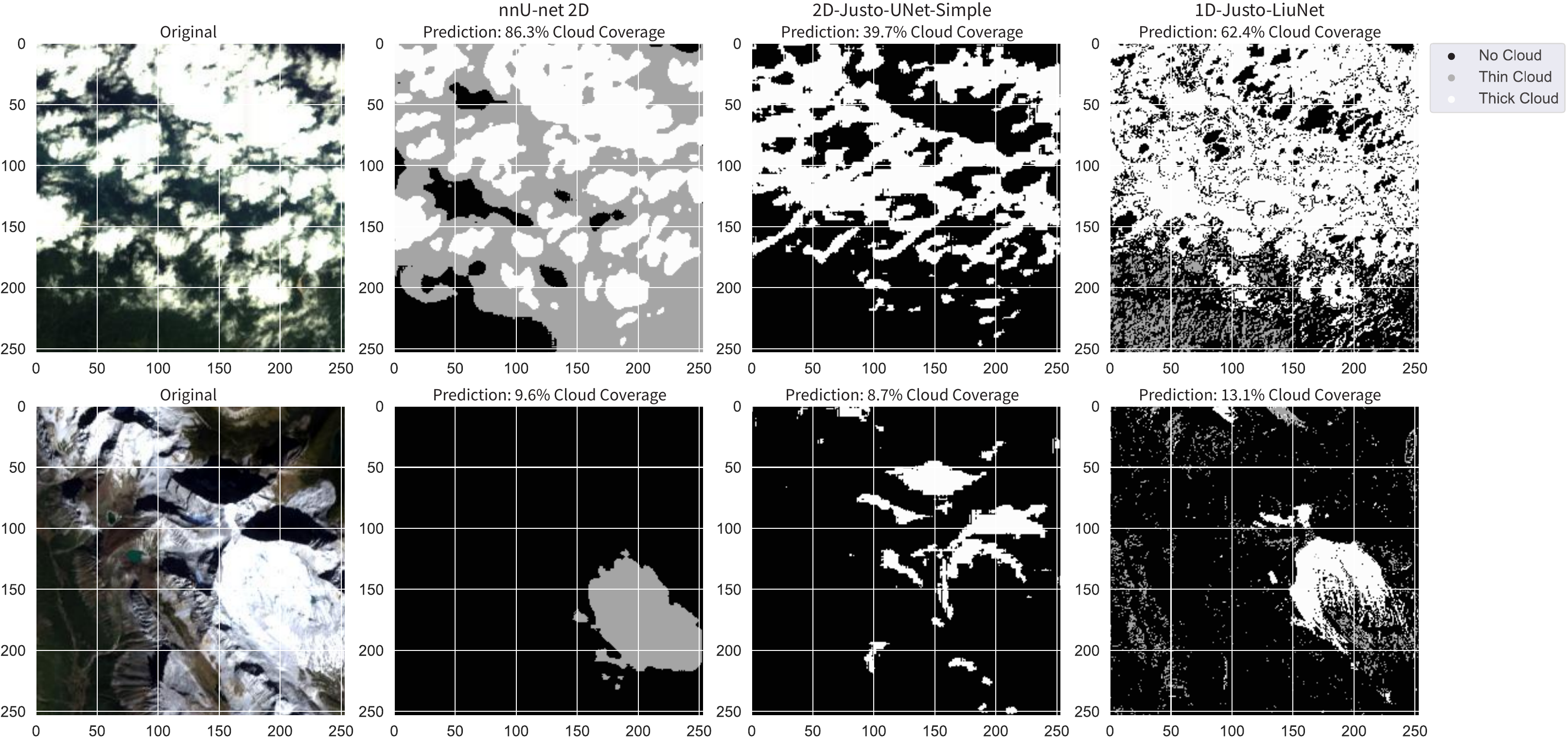}
\caption{Inference Results For Two Tiles From the Commercial Ziyuan-1 02 Dataset. The Top Tile Depicts Varying Cloud Thickness in an RGB Composite, While the Bottom One Includes a Snow-covered Mountainous Terrain. Next, the Predicted Masks Generated by the CNN Models for 6-channel Inference.}\label{Ziyuan}
\end{center}
\end{figure*}

In Table \ref{results}, we present the validation and testing performance for the three models, trained with different numbers of channels. For the test set, we highlight in bold the highest value for the metrics for each channel scenario. Additionally, we underline the top performance for each of the four metrics across all channel scenarios. Furthermore, in Table \ref{model_size_inf} we list the sizes of the models post-training, along with their speeds (inference times) to predict a new segmentation mask. These inference times are calculated as the average inference times of all tiles in both the validation and test sets. Finally, Fig. \ref{Ziyuan} illustrates a comparison of the inference results obtained from the three models on tiles extracted from the Ziyuan-1 02 dataset.

\begin{table}[!h]
\caption{Models' Performance Evaluation}  \label{results}
\centering
\resizebox{\columnwidth}{!}{%
\begin{threeparttable}
\begin{tabular}{crlrlrl}
\toprule

Number of channels                      &   \multicolumn{2}{c}{1} &   \multicolumn{2}{c}{6} &   \multicolumn{2}{c}{98} \\
Set                         &   Val.    & Test  &   Val.    & Test  &   Val.    & Test   \\ \hline
\multicolumn{1}{c}{}             &   \multicolumn{6}{c}{\textit{nnU-net 2D (PlainConvUNet)}}                                \\ \hline
PA     &   75.3    & \textbf{77.9}        &   80.4    & \textbf{80.2}        &   79.7    & 81.4         \\
DC           &  0.306    & \textbf{0.587}       &   0.599   & \underline{\textbf{0.607}}       &   0.605   & \textbf{0.582}        \\
CA &   90.9    & \textbf{87.8}        &   92.0    & 85.7        &   86.4    & 87.8         \\
CF1         &   0.92   & \textbf{0.864}       &   0.929   & 0.851       &   0.880   & 0.870        \\ \hline
\multicolumn{1}{c}{}             &   \multicolumn{6}{c}{\textit{2D-Justo-UNet-Simple}}                                     \\ \hline
PA     &   61.4    & 57.1          &   77.2    & 72          &   70.3    & 70.2         \\
DC           &   0.390   & 0.369       &   0.493   & 0.438       &   0.484   & 0.484        \\
CA &   76.1    & 75.5        &   87.5    & 87.8        &   77.3    & 85.7         \\
CF1         &   0.817   & 0.75        &   0.887   & \textbf{0.857}       &   0.773   & 0.811        \\ \hline
\multicolumn{1}{c}{}             &   \multicolumn{6}{c}{\textit{1D-Justo-LiuNet}}                                          \\ \hline
PA     &   65.6    & 64.8        &   76.7    & 74.0        &  84.3    & \underline{\textbf{82.2}}    \\
DC           &   0.435   & 0.397       &   0.472   & 0.443       &   0.572   & 0.548        \\
CA &   69.3    & 81.6        &   84.1    & \textbf{87.8}        &   90.9    & \underline{\textbf{98.0}}         \\
CF1          &   0.710   & 0.769       &   0.857   & 0.850       &   0.923   & \underline{\textbf{0.977}}    \\
\bottomrule
\end{tabular}
        \begin{tablenotes}
            \small
            \item \scriptsize{PA = Pixel Accuracy [\%]; DC = Dice Coefficient; CA = Classification Accuracy [\%];\\ CF1 = Classification F1 Score.}
        \end{tablenotes}
\end{threeparttable}
}
\end{table}

\begin{table}[!h]
\caption{Models' inference times and sizes}\label{model_size_inf}
\centering
\begin{tabular}{cccc}
\toprule
Number of channels                   & 1             & 6             & 98            \\ \hline
\multicolumn{1}{l}{}          & \multicolumn{3}{c}{\textit{nnU-net 2D (PlainConvUNet)}} \\ \hline
Model Size On Disk {[}MB{]}   & 255           & 256           & 257           \\
Model Size in Memory {[}MB{]} & 259           & 260           & 260           \\
Inference Time on CPU {[}s{]} & 0.889         & 0.859         & 1.078         \\ \hline
\multicolumn{1}{l}{}          & \multicolumn{3}{c}{\textit{2D-Justo-UNet-Simple}}      \\ \hline
Model Size On Disk {[}MB{]}   & \textbf{0.087}        & \textbf{0.090}         & 0.146         \\
Model Size in Memory {[}MB{]} & \textbf{0.016}         & \textbf{0.017}         & 0.036         \\
Inference Time on CPU {[}s{]} & \textbf{0.004}         & \textbf{0.005}         & \textbf{0.007}         \\ \hline
\multicolumn{1}{l}{}          & \multicolumn{3}{c}{\textit{1D-Justo-LiuNet}}           \\ \hline
Model Size On Disk {[}MB{]}   & 0.097         & 0.097         & \textbf{0.097}         \\
Model Size in Memory {[}MB{]} & 0.023         & 0.023         & \textbf{0.023}         \\
Inference Time on CPU {[}s{]} & 1,788         & 1.860         & 1.998         \\ \bottomrule
\end{tabular}
\end{table}

\section{Discussion}

Table \ref{results} shows that the best performance is achieved by 1D-Justo-LiuNet for the most extensive set of channels (98). Therefore, for the experimental EO-1 Hyperion dataset, this 1D-CNN-based model outperforms the tested 2D CNNs, aligning with previous findings for segmentation in HYPSO-1 data \cite{justo2023sea}, where the authors concluded that 1D CNNs can surpass 2D CNNs by concentrating on exploiting solely the extensive range of spectral channels within each image pixel, while disregarding the neighboring pixels. However, Table \ref{results} shows that reducing the number of channels diminishes the superiority of the 1D CNN over 2D CNNs due to the 1D-CNN's dependence on a vast array of spectral features for accurate predictions, also aligning with earlier research \cite{justo2023sea}. The table shows that when the number of channels decreases, 2D CNNs exhibit superior performance, with nnU-net surpassing 2D-Justo-UNet-Simple. Furthermore, we assume that by employing hyperparameter tuning (not feasible with the nnU-net network) and implementing data augmentation, we may arguably achieve higher classification accuracy, surpassing the current 88~\%. Moreover, the pair of 2D CNNs demonstrate similar or slightly better performance in scenarios with 6 channels compared to those with 98 channels. This highlights the efficacy of PCA analysis, indicating that in 2D models, a significant portion of HS data can be omitted while preserving the crucial information necessary for classifying cloudy tiles. Finally, Table \ref{results} shows that reducing the number of channels results in shorter inference times for the models. Additionally, 1D-Justo-LiuNet exhibits the slowest performance, followed by nnU-net, with 2D-Justo-UNet-Simple achieving significantly faster inference times. Furthermore, the lightweight 2D-Justo-UNet-Simple and 1D-Justo-LiuNet have comparable small model sizes, with a clear difference with nnU-net. For the 2D-Justo-UNet-Simple, the network shows inference times that are 172x shorter and a model size that is 2,844x smaller compared to the nnU-net model. This positions it as a highly promising option for space deployment due to its minimal memory usage. Similarly, 1D-Justo-LiuNet also shows promise due to its low memory usage, although it may necessitate hardware acceleration to mitigate its longer inference time. Finally, as depicted in Fig. \ref{Ziyuan}, when making predictions for the commercial sensor, the nnU-net model demonstrates subjectively the most effective cloud detection, followed by the lightweight 1D model, and then the lightweight 2D model. Additionally, it is noticeable that all models misclassify as clouds certain areas of the snow-covered mountains.

\section{Conclusion}

Our work investigated the latest CNN models for on-board cloud detection in hyperspectral satellite missions. Among the models, 1D-Justo-LiuNet (1D CNN) was found as the most accurate when tested on EO-1 Hyperion data, particularly when utilizing all the spectral channels. Along with its superior accuracy, it also achieved the most compact model size compared to other 2D-CNN-based models, yet at the cost of longer inference times. These results align with those from the HYPSO-1 mission, where the 1D CNN model demonstrated superior performance compared to 2D CNNs. Thus, we conclude that for maximizing accuracy in orbit, deploying 1D-Justo-LiuNet with extensive spectral channels is recommended. However, despite its efficient memory usage, the model suffers from longer inference times. Consequently, hardware acceleration, in fabric-logic within FPGAs, may be necessary to mitigate computational bottlenecks. Furthermore, reducing the number of channels diminishes the accuracy of the 1D CNN, with the 2D CNN nnU-net surpassing both 1D CNN and 2D-Justo-UNet-Simple. Thus, for fewer channels, 2D CNNs are preferable, especially nnU-net for its accuracy. However, nnU-net incurs higher memory usage and inference time compared to 2D-Justo-UNet-Simple. Consequently, while nnU-net excels for on-ground cloud detection, 2D-Justo-UNet-Simple is more suitable for in-orbit deployment due to its balanced accuracy-memory cost. Additionally, 2D-Justo-UNet-Simple's low inference time eliminates the need for hardware acceleration, unlike 1D-Justo-LiuNet. In short, for in-orbit cloud detection, the optimal balance between memory usage, inference times, and accuracy is achieved with 2D-Justo-UNet-Simple with 6 channels. Finally, as further work to improve the models' performance, we suggest expanding the training dataset, employing data augmentation or fine-tuning hyperparameters before model training.

\bibliographystyle{IEEEtran}
\bibliography{bib_database}



\end{document}